\documentclass[10pt,twocolumn,letterpaper]{article}

\usepackage{cvpr}              % To produce the CAMERA-READY version
\usepackage{xcolor}
\usepackage{wasysym} 
\usepackage{pifont} 
\usepackage{booktabs}
\usepackage{graphicx}
\usepackage{multirow}
\usepackage{amsmath}
\usepackage{caption}

\usepackage{tikzsymbols}

\definecolor{goodgreen}{RGB}{46, 204, 113}  
\definecolor{normalyellow}{RGB}{241, 196, 15}
\definecolor{badred}{RGB}{231, 76, 60}    

\newcommand{\good}{\Smiley[1.2][goodgreen!60!white]}
\newcommand{\bad}{\Sadey[1.2][badred!60!white]}
\newcommand{\normal}{\Neutrey[1.2][normalyellow!80!orange]}
\definecolor{cvprblue}{rgb}{0.21,0.49,0.74}
\usepackage[pagebackref,breaklinks,colorlinks,allcolors=cvprblue]{hyperref}

\def\paperID{7420} % *** Enter the Paper ID here
\def\confName{CVPR}
\def\confYear{2026}

\title{WorldStereo: Bridging Camera-Guided Video Generation and Scene Reconstruction via 3D Geometric Memories}

\author{Yisu Zhang$^{1,2}$\footnotemark[1], ~ Chenjie Cao$^{2}$\footnotemark[1], ~ Tengfei Wang$^{2}$\footnotemark[2], 
\\ Xuhui Zuo$^{2}$, ~ Junta Wu$^{2}$, ~ Jianke Zhu$^{1}$\footnotemark[3], ~ Chunchao Guo$^{2}$ \\
\small $^{1}$Zhejiang University ~~$^2$ Tencent Hunyuan \\
}

\begin{document}

\twocolumn[{
\renewcommand\twocolumn[1][]{#1}
\maketitle
\begin{center}
\includegraphics[width=1\linewidth]{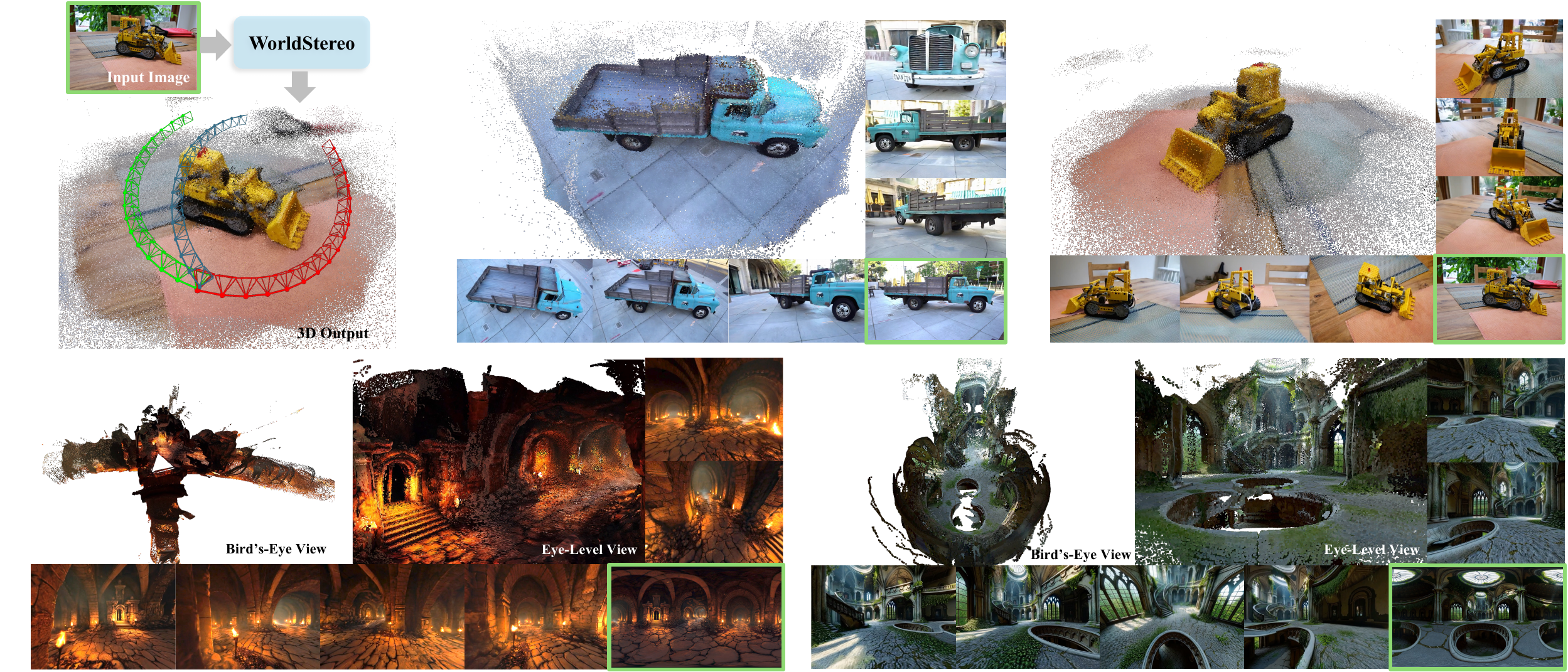}
\vspace{-0.2in}
\captionof{figure}{ 
\textbf{WorldStereo enables high-quality 3D scene generation based on single-view or panoramic inputs.} The input reference views are framed in green. We present point clouds reconstructed from videos generated by WorldStereo: the top two perspective scenes use WorldMirror~\cite{liu2025worldmirror}, while the bottom two panoramic scenes are aligned via monocular depth maps~\cite{wang2025moge}. 
\label{fig:teaser}}
\end{center}
}]

\renewcommand{\thefootnote}{\fnsymbol{footnote}}
\footnotetext[1]{Equal Contribution.  \footnotemark[2]Project Lead.
\footnotemark[3]Corresponding author}

% \begin{figure*}[!t]
%     \centering
%     \includegraphics[width=1\linewidth]{fig/teaser.pdf}
%     \caption{\textbf{WorldStereo enables high-quality 3D scene generation.}}
%     \label{fig:teaser}
% \end{figure*}

\begin{abstract}
Recent advances in foundational Video Diffusion Models (VDMs) have yielded significant progress. Yet, despite the remarkable visual quality of generated videos, reconstructing consistent 3D scenes from these outputs remains challenging, due to limited camera controllability and inconsistent generated content when viewed from distinct camera trajectories.
In this paper, we propose \textbf{WorldStereo}, a novel framework that bridges camera-guided video generation and 3D reconstruction via two dedicated geometric memory modules. 
Formally, the global-geometric memory enables precise camera control while injecting coarse structural priors through incrementally updated point clouds.
Moreover, the spatial-stereo memory constrains the model's attention receptive fields with 3D correspondence to focus on fine-grained details from the memory bank.
These components enable WorldStereo to generate multi-view-consistent videos under precise camera control, facilitating high-quality 3D reconstruction.
Furthermore, the flexible control branch-based WorldStereo shows impressive efficiency, benefiting from the distribution matching distilled VDM backbone without joint training.
Extensive experiments across both camera-guided video generation and 3D reconstruction benchmarks demonstrate the effectiveness of our approach. Notably, we show that WorldStereo acts as a powerful world model, tackling diverse scene generation tasks (whether starting from perspective or panoramic images) with high-fidelity 3D results. Models will be released on https://github.com/FuchengSu/WorldStereo.
\end{abstract}

\vspace{-0.1in}
\section{Introduction}
\label{sec:intro}

Recently, remarkably developed Video Diffusion Models (VDMs)~\cite{Blattmann2023svd,yang2025cogvideox,Tim2024sora,kling,runwaygen3,kong2024hunyuanvideo,wang2025wan} have made impressive advances, achieving remarkable performance in photorealistic video synthesis.
These models have showcased significant potential across a broad range of applications, including virtual reality, digital content creation, and embodied AI.
Meanwhile, camera-controllable VDMs have also enabled substantial progress, incorporating various camera and action controls~\cite{He2024Cameractrl,bahmani2024ac3d,zheng2024cami2v,he2025cameractrl,liang2024wonderland}, as well as leveraging explicit guidance like point clouds~\cite{yu2024viewcrafter,feng2024i2vcontrol,ma2025you,ren2025gen3c,li2025realcam,popov2025camctrl3d,cao2025uni3c}, optical flow~\cite{jin2025flovd,burgert2025go}, and tracking points~\cite{gu2025diffusion,wang2025ati}.

Despite these breakthroughs, current camera-guided VDMs remain limited in recovering consistent and reliable 3D scene reconstructions, even when paired with advanced feed-forward 3D reconstruction approaches~\cite{yang2025fast3r,wang2025continuous,wang2025vggt,keetha2025mapanything}, failing to step toward generalizable world models. 
A fundamental challenge lies in enabling VDMs to \emph{generate videos that capture sufficiently diverse and comprehensive viewpoints of the target scene, while maintaining precise camera control and high-fidelity visual quality}.
Existing camera-guided VDMs struggle to preserve consistency across varied camera trajectories, leading to ambiguous and blurry reconstructions. Intuitively, extending them into longer sequences can capture richer viewpoint coverage with natural consistency from the global attention of diffusion transformers (DiTs)~\cite{peebles2023scalable}, but this often comes at inferior video quality and prohibitive computation for both fine-tuning and inference. 
On the other hand, autoregressive (AR) VDMs~\cite{chen2024diffusion,huang2025self} improved the efficiency for long video generation via sequential synthesis, yet they suffer from limited camera precision~\cite{song2025generative} and error accumulation~\cite{huang2025self}. 
Moreover, both long video generation and AR models lack mature open-source communities compared to regular VDMs like HunyuanVideo~\cite{kong2024hunyuanvideo}, CogVideoX~\cite{yang2025cogvideox}, and Wan~\cite{wang2025wan}.

\begin{table}
\centering
\caption{\textbf{Different video generation schemes for 3D reconstruction.} \textbf{Long-Bi} VDMs produce long trajectories in a single pass to cover diverse viewpoints. \textbf{AR} models sequentially generate long videos in an autoregressive manner. \textbf{Multi-Bi-Mem} (ours) achieves multiple consistent generations based on a powerful open-released VDM~\cite{wang2025wan} with complementary viewpoints and memory mechanisms for integrated reconstruction.}
\vspace{-0.1in}
\label{tab:teaser_comparison}
\resizebox{\linewidth}{!}{
\fontsize{7pt}{8pt}\selectfont
\setlength{\tabcolsep}{2.5pt}
\begin{tabular}{lccc}
\toprule
\textbf{Paradigms} & \textbf{Long-Bi} & \textbf{AR} & \textbf{Multi-Bi-Mem} \\
\midrule
Receptive Field & Bidirectional & Autoregressive & Bidirectional \\
Trajectory Length/Num & Long/Single & Long/Single & Medium/Multiple \\
\midrule
Consistency & \normal & \normal & \good \\
Precise Camera Control & \good & \normal & \good \\
Video Quality & \normal & \bad & \good \\
Efficiency & \bad & \good & \good \\
Community Progress  & \bad & \bad & \good \\
\bottomrule
\end{tabular}}
\vspace{-0.1in}
\end{table}

% high-level
To overcome these challenges, we present \textbf{WorldStereo}, a novel framework that bridges the gap between camera-guided VDMs and 3D scene reconstruction by enabling \emph{consistent multi-trajectory video generations with geometry-aware memories}, as summarized in \Cref{tab:teaser_comparison}.
Specifically, two complementary mechanisms, \emph{global-geometric} and \emph{spatial-stereo} memory, are incorporated into off-the-shelf VDMs to memorize coarse structures and fine-grained details, respectively. 
These two components enable precise and coherent video synthesis across diverse trajectories. 
WorldStereo subtly sidesteps the long sequence generation and largely preserves the generalization and usability of pre-trained VDMs, resulting in impressive 3D reconstruction as shown in \Cref{fig:teaser}.

% in detail
Formally, WorldStereo is built upon the camera-guided VDM, Uni3C~\cite{cao2025uni3c}, with the extended capability to memorize consistent generation along different trajectories.
We first augment Uni3C's point cloud guidance as an incrementally updated Global-Geometric Memory (GGM) through iterative feed-forward 3D reconstruction. 
However, we find that GGM fails to preserve fine-grained details. Inspired by stereo matching~\cite{marr1976cooperative}, we introduce the Spatial-Stereo Memory (SSM): learning spatial coherence between the generated novel views and retrieved views from a deliberate memory bank by establishing explicit 3D correspondences.
Furthermore, we constrain the attention receptive fields of each novel view to focus on the specific retrieved one, thereby enhancing detailed consistency.
Additionally, WorldStereo inherits the flexibility of Uni3C: all pixel-wise aligned conditions are injected via the ControlNet branch, which is highly compatible with the distribution matching distillation (DMD)~\cite{yin2024one,yin2024improved} and eliminates the need for joint training.
This technique significantly accelerates WorldStereo under 4-step inference without compromising generalization and consistency.

Extensive experiments confirm the effectiveness of WorldStereo, including both in-domain and out-of-distribution benchmarks. Our approach achieves superior camera-motion accuracy and higher-quality video generation. 
To further demonstrate its contributions to 3D reconstruction, we present a new 3D reconstruction benchmark to evaluate the output quality of camera-guided VDMs.
We meticulously process and crop the ground-truth point clouds for Tanks-and-Temples~\cite{knapitsch2017tanks} and MipNeRF360~\cite{barron2022mip}, enabling comprehensive assessment of 3D consistency, visual quality, and camera trajectory fidelity.
Our main contributions are summarized as follows:
\begin{itemize}
    \item \textbf{Geometry-aware memory enhanced VDM.} WorldStereo contains two complementary memory mechanisms to generate multi-trajectory consistent videos tailored for 3D reconstruction.
    \item \textbf{Flexible framework with efficient inference.} Leveraging pixel-wise aligned ControlNet injection, WorldStereo can be accelerated via DMD without requiring joint training or sacrificing generalization.
    \item \textbf{Customized evaluation.} We present a new 3D reconstruction benchmark to evaluate the outcome quality of camera-guided VDMs.
\end{itemize}
% \begin{itemize}
%     \item \textbf{Geometry-aware memory based VDM.} We propose a novel and lightweight module that injects a geometric prior to efficiently generate videos with high geometric fidelity. This is achieved by training only a small set of parameters, without requiring full fine-tuning of the base model.
%     \item \textbf{Efficient Inference.} The flexible control-branch-based framework demonstrates remarkable efficiency, due to a distilled VDM backbone matched to the distribution and requiring no joint training.
%     \item \textbf{A 3D Benchmark for Geometric Consistency.} We introduce a new benchmark for the quantitative evaluation of multi-view consistency from a 3D perspective. Using this benchmark, we demonstrate that our method can effectively recover complete and accurate scene geometry from a single image.
% \end{itemize}

\section{Related Work}
\label{sec:related_work}

\noindent\textbf{Camera-Guided Video Generation.}
Taming high fidelity video generations~\cite{xing2024dynamicrafter,zheng2024open,lin2024open,kong2024hunyuanvideo,yang2025cogvideox,wang2025wan} under controllable viewpoints and camera trajectories serves as a pivotal pathway for world simulation.
Some works tried to implicitly control the VDM under specific camera motion via tailored LoRA tuning~\cite{guo2023animatediff,Blattmann2023,sun2024dimensionx,zhang2025lion}. Meanwhile, implicitly discrete controls are also widely adopted in action-based VDMs~\cite{zhang2025matrix,li2025hunyuan} and AR models~\cite{decart2024oasis,guo2025mineworld,he2025matrix}.
MotionCtrl~\cite{wang2024motionctrl} first explicitly injected camera poses into the pre-trained VDM. Subsequent works further extend the camera presentation into Plücker ray~\cite{He2024Cameractrl,li2025realcam, yang2024direct,bahmani2024ac3d,he2025cameractrl}.
To enhance the camera control precision under metric scales, point clouds~\cite{yu2024viewcrafter,feng2024i2vcontrol,ma2025you,ren2025gen3c,li2025realcam,popov2025camctrl3d,cao2025uni3c}, mesh~\cite{hu2025ex,yang2025matrix}, optical flow~\cite{jin2025flovd,burgert2025go}, and tracking points~\cite{gu2025diffusion,wang2025ati} have been employed for camera-guided VDMs, spanning both training-based and training-free manners~\cite{hu2024motionmaster,hou2024training,zhang2025recapture}.
Despite the precise camera controllability achieved by these pioneering efforts, we note that they struggle to produce convincing 3D reconstruction due to constrained video lengths and inadequate consistency caused by memoryless visual conflicts.

\noindent\textbf{Memory-based Video Generation.}
Existing video generation fails to maintain geometrically accurate structures during long-trajectory synthesis. 
A straightforward solution involves training VDMs with extended context lengths~\cite{song2025history,valevski2024diffusion,chen2025skyreels} to memorize more visual clues, yet computationally costly.
To reduce the of long contexts computation, concurrent works either compress prior frames into a condensed context window~\cite{gu2025long,zhang2025packing} or inject historical frames via attention mechanisms~\cite{zhou2025stable,schneider2025worldexplorer,xiao2025worldmem}, which inevitably cause information loss and compromise 3D consistency. 
Another research direction iteratively reconstructs 3D representations from historical frames, utilizing these as either conditional guidance~\cite{ma2025you,yu2024viewcrafter,ren2025gen3c,gu2025diffusion,wu2025video} or retrieved references~\cite{yu2025context,li2025vmem} for future contexts synthesis. But they also suffer from error accumulation and degraded fine-grained details.
Our geometry-aware memory preserves both 3D consistency and high-fidelity details by unifying 3D correspondence modeling and a tailored attention mechanism.

% \paragraph{Scene Reconstruction from Single View.}
\noindent\textbf{Feed-Forward 3D Reconstruction.}
In contrast to traditional Structure from Motion (SfM)~\cite{galliani2015massively,schoenberger2016mvs,pan2024glomap}, Multi-View Stereo (MVS)~\cite{yao2018mvsnet,gu2020cascade,zhang2023vis,cao2024mvsformer++}, and learning-based SLAM~\cite{li2025megasam,huang2025vipe}, recent feed-forward 3D reconstruction approaches efficiently predict both camera poses, point clouds, and depth maps within a single forward pass.
Dust3R~\cite{wang2024dust3r} pioneered learning two-view correspondence through powerful transformer models pre-trained by cross-view inpainting~\cite{weinzaepfel2023croco}. 
Many follow-ups further investigated this realm, including  multi-view-based~\cite{yang2025fast3r,yang2025fast3r}, sequential-based~\cite{chen2025long3r,wang2025continuous}, dynamic-based~\cite{zhang2024monst3r,lu2025align3r} and conditional priors-based~\cite{jang2025pow3r,keetha2025mapanything,liu2025worldmirror} reconstructions.
For impressive visual reconstruction, some works have focused on feed-forward 3D Gaussian Splatting (3DGS) generation~\cite{chen2024mvsplat}, enabling fast, optimization-free 3DGS representations.
However, 3D reconstruction methods follow ``what you see is what you get'', \ie, they require sufficient images to capture 3D structures, suffering from inferior results explored beyond observed viewpoints.

% \TODO{update more citations here.}

\noindent\textbf{3D Scene Generation.}
One popular scene generation workflow involves iterative \emph{warp-and-inpaint}, combining depth estimation, alignment, and image inpainting~\cite{fridman2023scenescape,hollein2023text2room,liang2024luciddreamer,shriram2024realmdreamer,yu2025wonderworld,wang2025vistadream}. While iterative scene generations perform well in novel view synthesis (NVS), they suffer from prohibitive per-scene optimization and inconsistent geometry across views. 
Another prominent paradigm can be summarized as \emph{generate first, then reconstruct}, yielding various 3D outcomes from multi-view images~\cite{gao2024catd,cao2025mvgenmaster,zhou2025stable}, panoramas~\cite{hunyuanworld2025tencent} and camera-guided videos~\cite{sun2024dimensionx,yu2024viewcrafter,zhao2024genxd,liang2024wonderland}. 
However, both of them face distinct limitations: multi-view images cover adequate viewpoints but lack view consistency, while generated videos are insufficiently long to capture disparate viewpoints.
Recent works have unified generation and reconstruction in end-to-end frameworks, simultaneously modeling depth~\cite{huang2025voyager,zhu2025aether,chen2025deepverse}, 3DGS~\cite{yang2024direct,bahmani2025lyra,yang2025flash}, and point map~\cite{szymanowicz2025bolt3d,zhang2025world} rather than relying solely on RGB frames.
Though these methods largely alleviate 3D consistency issues, they are highly data-hungry and require substantial training, potentially hindering the generalization of foundation models. 
Our method preserves the original output formulation of VDMs with retained generalization, while capturing more consistent viewpoints via a novel memory mechanism for 3D reconstruction.

\section{Method}

\begin{figure*}[h]
    \centering
    \includegraphics[width=1.0\linewidth]{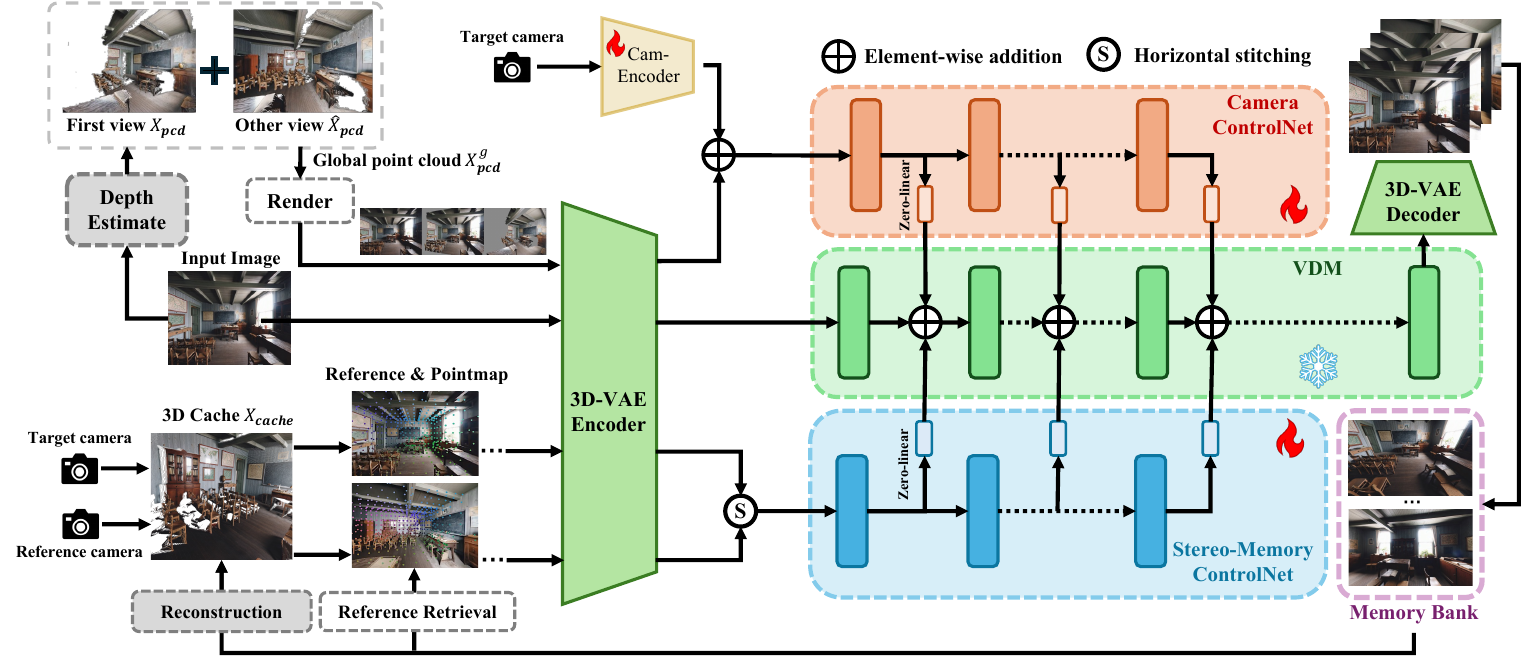}
    \vspace{-0.15in}
    \caption{\textbf{Overview of WorldStereo.} 
    WorldStereo comprises two ControlNet branches. The camera branch ensures precise camera control and Global-Geometric Memory (GGM), depending on global point clouds; the Spatial-Stereo Memory (SSM) branch leverages retrieved reference frames and pointmap (3D correspondence) guidance obtained from the 3D cache to further preserve fine-grained consistency. We omit the diffusion noise part for simplicity.}
    \label{fig:framework}
    \vspace{-0.15in}
\end{figure*}

\noindent\textbf{Overview.}
Given a single image, WorldStereo aims to reconstruct a complete 3D scene following the \emph{``generate first, then reconstruct''} paradigm. 
We summarize the generation pipeline in \Cref{fig:framework}. 
We first introduce the camera-guided VDM with point cloud guidance and our basic memory storage in \Cref{sec:perliminary}. 
Next, the global-geometric (\Cref{sec:global_geo_memory}) and spatial-stereo memories (\Cref{sec:spatial_stereo_memory}) are proposed to generate multiple video sequences with consistent 3D geometry and textures for subsequent 3D reconstruction. To improve inference efficiency, WorldStereo adopts a modified distribution matching distillation (\Cref{sec:dmd}).
% achieves a 20 times speedup without compromising generation quality by applying
% Finally, we establish a new benchmark for evaluating the geometric fidelity of generative models, enabling a systematic comparison of 3D consistency across methods when provided with a single monocular input (Sec. 3.4).

\subsection{Preliminaries}
\label{sec:perliminary}

% Diffusion models comprise two cascaded Markov processes: a forward diffusion process and a reverse generative process. Building on this framework, latent video diffusion model (VDM) that employs a causal 3D VAE for spatio-temporal compression and a Diffusion Transformer (DiT)~\cite{} to model the latent distribution. 

\textbf{Uni3C}~\cite{cao2025uni3c} is used as the base camera-guided VDM in WorldStereo. 
Formally, Uni3C is built upon the pre-trained Wan-I2V~\cite{wang2025wan}, integrated with a lightweight ControlNet branch~\cite{zhang2023adding} trained from scratch. 
Except for the camera Plücker rays, Uni3C incorporates point clouds extracted from the reference frame as 3D geometric priors, derived via back-projected monocular depth. Point clouds $X_{pcd}$ of the reference image can be presented as:
\begin{equation}
X_{pcd}(x) \simeq R_{c \rightarrow w} D(x) K^{-1} \hat{x},
\label{eq_pcd}
\end{equation}
where $R_{c \rightarrow w}$ denotes the camera-to-world pose matrix; $D(\cdot)$ is the depth at pixel $x$ estimated by MoGe~\cite{wang2025moge}; $K^{-1}$ is the inverse of the camera intrinsic matrix; $\hat{x}$ is the homogeneous pixel coordinate.
% Given conditional text $c_{txt}$, reference image $c_{img}$, and $N$ target camera viewpoints $\{P^i\}_{i=1}^N$, Uni3C models the distribution of the diffusion velocity $v_{\theta}$ as $p(v|c_{txt},c_{img},c_{pcd},P)$, where $c_{pcd}$ indicates the guiding frames rendered from point clouds $X_{pcd}$.
Point clouds serve as strong geometric guidance to facilitate fast convergence and precise camera control, avoiding compromising the generalization of frozen foundational VDMs.
% By rendering this point cloud from the target camera trajectory, we provide a strong geometric guidance to the ControlNet. A key advantage of this design is that it enables precise control over complex camera trajectories by only fine-tuning the ControlNet module, while keeping the pretrained VDM frozen.

\noindent\textbf{Memory Bank \& 3D Cache.}
WorldStereo incorporates two memory components: a 2D memory bank and a 3D cache, as in \Cref{fig:framework}.
Formally, generated video frames are first temporally downsampled and stored in the memory bank as $\{I_{mem}\}_{m=0}^{M}$, serving to retrieve spatially similar reference views for the subsequent spatial-stereo memory.
Note that the initial condition image and perspective views split from the 360$^{\circ}$ panorama are also included in the memory bank.
The 3D cache saves global point cloud set $X_{cache}$, reconstructed based on memory bank images using the feed-forward 3D reconstruction method: WorldMirror~\cite{liu2025worldmirror}. 
Specifically, we incrementally reconstruct point clouds for each generated video. To tackle long sequences, disparate 3D caches are merged by aligning point clouds from overlapping views via Umeyama transformation~\cite{umeyama1991least}.

\subsection{Global-Geometric Memory}
\label{sec:global_geo_memory}

We propose Global-Geometric Memory (GGM) that iteratively updates point cloud conditions, serving as global 3D priors for generating multiple consistent videos, iterative video continuation, and even supporting panorama-based 3D generation, as verified in \Cref{sec:qualitative}.
As discussed in~\cite{cao2025uni3c}, point clouds just provide camera guidance rather than forcing the VDM to overfit to these 3D presentations.
While this phenomenon is reasonable for camera-guided VDMs to preserve overall generalization from the degradation caused by inferior monocular depth, it causes our model to ignore most geometric structures brought by point clouds, even when the point clouds themselves are perfectly reconstructed.

To overcome this, we fine-tune the original control branch of WorldStereo using extended global point clouds $X^g_{pcd}$ beyond the reference points ($X_{pcd}$ from Eq.~\ref{eq_pcd}) as:
\begin{equation}
X^g_{pcd}=[X_{pcd}, \hat{X}_{pcd}],
\label{eq_global_pcd}
\end{equation}
where $\hat{X}_{pcd}$ denotes point clouds from other views. 
To avoid overfitting novel views' point clouds during training, we introduce a point cloud masking strategy: randomly dropping a subset of points from target views to confirm robustness to partially geometric missing. We detail the sampling strategy in supplementary.
For inference, we use the incrementally updated 3D cache as $\hat{X}_{pcd}$, which is aligned to the same coordinate as $X_{pcd}$ via the Umeyama transformation~\cite{umeyama1991least} among the overlapping point clouds from monocular depth estimation and WorldMirror.
Notably, GGM is also compatible with panorama-based 3D generation. 
Since panoramas capture 360$^{\circ}$ views, VDMs have to maintain visually consistent content during viewpoint transitions.
To this end, we follow the MoGe panorama depth estimation~\cite{wang2025moge} to construct panoramic point clouds as the initialized 3D cache for our panoramic 3D generation.
% This procedure acts as a form of data augmentation, forcing the model to learn a more robust and continuous representation of the scene's geometry rather than memorizing the sparse point distributions from specific viewpoints. 
% which uses only the initial point cloud as guidance for camera motion, we also treat the point cloud as a memory reference. For example, when point-cloud cues exist outside the initial camera view, WorldStereo can faithfully reflect these geometric hints in the generated video once the camera moves into those regions. This is especially beneficial when the input view is merely a crop of a panorama: WorldStereo memorizes the additional panoramic point cloud and preserves consistency with the panorama as the camera transitions to novel viewpoints.
% Specifically, the input condition $X_{global}$ is defined as 
% $
% X_{global} = X_{pcd} \oplus X_{add} ,
% $
% where $X_{pcd}$ is point cloud obtained from eq.~\eqref{eq1}, $X_{add}$ is an optional auxiliary spatial point-cloud condition that may be empty and $\oplus$ is the concatenation operation. As shown in \TODO{Fig.2}, WorldStereo captures global point-cloud information as memory, thereby maintaining consistency with the panorama under new viewpoints.

\begin{figure}
    \centering
    \includegraphics[width=1.0\linewidth]{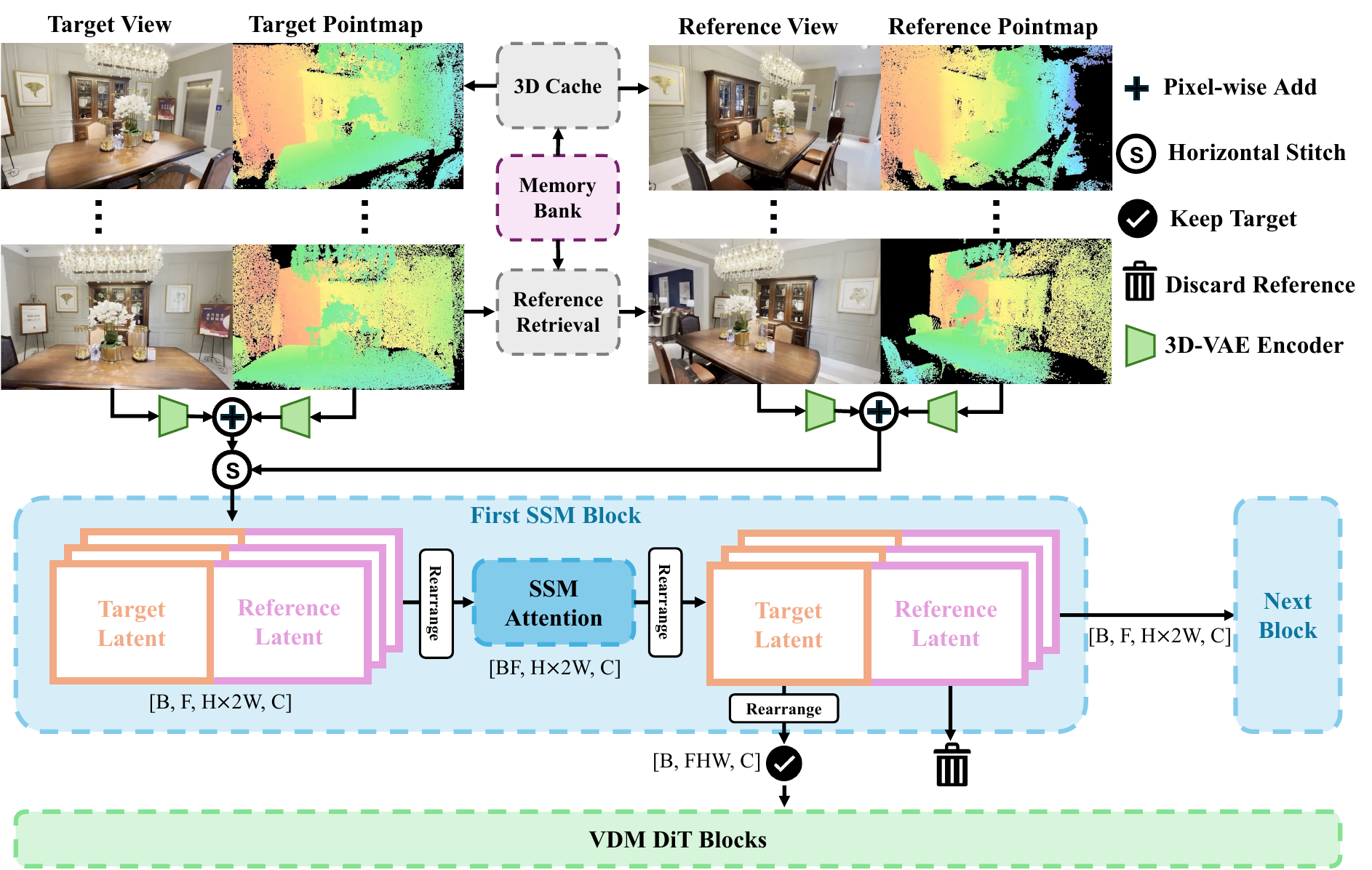}
    \vspace{-0.2in}
    \caption{\textbf{Spatial-Stereo Memory (SSM).} Reference views are retrieved from the memory bank, while pointmaps for both target and reference views are constructed based on the 3D cache. In SSM attention, we horizontally stitch each target-reference pair and rearrange the tensor shape to make each target frame's features focus on the specifically retrieved reference. B, F, H, W, C indicate dimensions of batch, frame, height, width, and channels.\label{fig:stereomem}}
    \vspace{-0.2in}
\end{figure}

\subsection{Spatial-Stereo Memory}
\label{sec:spatial_stereo_memory} 

Although GGM maintains coarse structures using point clouds from 3D cache, it struggles to preserve fine-grained details, as shown in~\Cref{fig:ablation_study}.
Many previous studies~\cite{yu2025context,zhou2025stable,schneider2025worldexplorer,li2025vmem} retrieve historical reference frames and jointly model all frames via full-attention. However, this formulation requires substantial post-training to enable VDMs to adapt to long sequences.
Moreover, we cannot guarantee the continuity of retrieved frames (\eg, panoramic scenarios). These disparate, unordered reference views further hinder the VDM learning process.
Thus, we get inspirations from the traditional stereo matching~\cite{marr1976cooperative} and the reference-based inpainting~\cite{cao2024leftrefill} and propose the Spatial-Stereo Memory (SSM) as illustrated in \Cref{fig:stereomem}.
Formally, we discretely retrieve reference views and spatially stitch each with its corresponding target view. We then constrain the attention receptive field to each reference-target pair, facilitating enhanced fine-grained detail recovery.

In practice, given $N$ target poses, we first sample $F=N/4$ poses uniformly and retrieve their nearest-neighbor frames from the memory bank as references.
We extend the retrieval strategy of~\cite{yu2025context} from 2D planes to 3D spaces, selecting views where the volumetric overlapping fields-of-view (FoV) between the target and reference camera frustums are maximized.
To ensure that fine-grained details are preserved after 3D-VAE encoding, we separately encode each retrieved reference view as an independent image, yielding latent features $\{z_{ref}\}_{i=1}^{F}$.
Then we horizontally stitch target and reference latent features as $z_{stitch}=[z_{tar};z_{ref}]\in\mathbb{R}^{F\times 2HW\times C}$, where $H,W,C$ denote the height, width, and channel dimension of latent features.
To improve the geometry-aware perception, we incorporate a \textit{pointmap} to each stitched target-reference pair $z_{stitch}$, which indicates 3D corresponding information derived from our 3D cache.
Specifically, the pointmap records point cloud positions of the target-reference pair in the 3D world coordinate. We normalize and colorize this into an RGB format, which is then encoded into latent features by 3D-VAE---denoted as $\hat{z}_{tar}$ and $\hat{z}_{ref}$ for target and reference pointmaps, respectively. 
The stitched pointmap latents are thus formulated as $\hat{z}_{pm}=[\hat{z}_{tar};\hat{z}_{ref}]\in\mathbb{R}^{F\times 2HW\times C}$. 
Subsequently, we achieve the final inputs for the SSM branch as $z_{ssm}=z_{stitch}+\hat{z}_{pm}$. Our ablation studies in~\Cref{fig:ablation_study}(c)(d) confirm that this 3D correspondence information is critical to SSM's performance.
The model architecture of the SSM branch is similar to the camera branch, comprising 20-layer DiT blocks trained from scratch.
A key distinction lies in the SSM attention constraint: instead of full attention, we limit the attention receptive field such that each target-reference pair attends exclusively to its own features---facilitating more precise fine-grained learning.
As shown in \Cref{fig:stereomem}, we rearrange the feature to $[BF,H*2W,C]$ and restrict SSM attention to operate solely along the $H*2W$ dimension. Then, only the target features will be added to the main VDM block.

\paragraph{Data Curation.}
Intuitively, training SSM needs multi-view videos to construct reference-target pairs, which is difficult to achieve in the real world. 
To overcome this, we generate training pairs by temporally misaligned sampling existing multi-view data~\cite {ling2024dl3dv,xia2024rgbd,tartanair2020iros,arnold2022map}, ensuring that the reference video and target video have a temporal overlap of 30\% to 90\%. 
Then we randomly shuffle and apply masks to reference views to confirm the robustness, as well as simulate the disorder and discreteness of real-world retrieval scenarios. More details are discussed in the supplementary.

\subsection{Acceleration via DMD}
\label{sec:dmd}

We apply the modified Distribution Matching Distillation (DMD)~\cite{yin2024improved} to accelerate the inference of WorldStereo. DMD extends the idea of Variational Score Distillation (VSD)~\cite{wang2023prolificdreamer}, distilling a few-step diffusion student $G_\theta$ through the approximate Kullback-Liebler (KL) divergence built from the difference between the frozen real score function $s_{real}$ and the trainable fake score function $s_{fake}$.
The update gradient of DMD can be written as:
\begin{equation}
\hspace*{-1em}
\nabla \mathcal{L}_{\text{DMD}} = -\underset{t}{\mathbb{E}} \left( \int \left( s_{\text{real}}(x_t, t) - s_{\text{fake}}(x_t, t) \right) \frac{dx_t}{d\theta} dz \right),
\label{eq:dmd1}
\end{equation}
where $x=G_\theta(z)$ denotes the student generation given random Gaussian noise $z\sim\mathcal{N}(0;\mathbf{I})$ and $t\sim\mathcal{U}(0,1)$, while $x_t\sim q_t(x_t|x,t)$ indicates the forward diffusion process.

The generator $G_\theta$ of WorldStereo is distilled into a 4-step DiT.
$G_\theta$, $s_{real}$, $s_{fake}$ are all initialized from the camera-guided VDM (Uni3C): $s_{real}$ is frozen, while $G_\theta$ and $s_{fake}$ are trainable. 
Following~\cite{yin2024improved}, we train $s_{fake}$ 5 times per generator update. The stochastic gradient truncation~\cite{huang2025self} is employed to stabilize the training phase. We omit the GAN loss, as we found its impact to be insignificant while substantially slowing down training.
Moreover, the DMD training is based on pure camera-guided video generation without any memory training.
To decouple the control and the basic few-step generation capabilities, we freeze the camera-guided control branch of the generator $G_\theta$, leaving only its main backbone trainable.
Notably, both the camera-guided and memory-based control branches can be generalized to the distilled generator $G_\theta$ without any joint fine-tuning. This largely simplifies the DMD training process while preserving the generalization (annotated memory data requires well-aligned depth maps, which is less than the camera control data).
Importantly, all parameters of the fake score function remain trainable, facilitating $s_{fake}$ to track the generator's distribution.
Additionally, we empirically find that retaining high-quality and relatively easy trajectories is critical for the stable training of DMD. 
The student model prefers to attend to some artifacts (\eg, over-saturation and hallucination) from the teacher model, especially under challenging trajectories or low-quality reference images.
As verified in \Cref{tab:ood_exp3}, this data filter strategy does not degrade the camera controllability.

\section{Experiments}

\subsection{Implementation Details}
\label{sec:impl_detail}

The frozen VDM of WorldStereo is built upon Wan2.1-14B-I2V~\cite{wang2025wan}, with the camera ControlNet retrained under the same setting as~\cite{cao2025uni3c} for 8,000 steps under a batch size of 32. 
For the GGM training in the second phase, we fine-tune the camera ControlNet with the global point cloud augmentation for 4,000 steps.
For the third phase of SSM training, we train a new branch from scratch for 6,000 steps based on tailored memory-retrieval data as detailed in \Cref{sec:spatial_stereo_memory}.
Training both memory mechanisms costs 60 hours on 64 NVIDIA H20 GPUs.
For DMD training, we set the classifier-free guidance (CFG) scale to 5.0 for the real score function, leading to a CFG-free generator with 2x inference efficiency. Furthermore, we reduce the inference denoising steps from 40 to 4, achieving an overall speedup of 20×.
We train DMD for 1,000 steps within 13 hours under the same resources.
All training data are resized to 480p with flexible aspect ratios. We also verify that our method can be generalized to 720p inference in supplementary. 

% We use multi-resolution images scaled form [480 $\times$ 768, 512 $\times$ 720, 576 $\times$ 640, 608 $\times$ 608, 640 $\times$ 576, 720 $\times$ 512, 768 $\times$ 480] of 81 frames to finetune. When finetuning the camera ControlNet, the learning rate is warmed up to 1e-5 for 200 steps and then fixed, whereas for training the stereo-memory ControlNet from scratch, the learning rate is warmed up to 1e-5 over 400 steps.

% We finetune the camera ControlNet for 4000 steps with a batch size of 16 and train the stereo-memory ControlNet for 6000 steps with a batch size of 16. The total time of training both branches with 64 NVIDIA H20 GPUs for 60 hours. Following the previous work, we randomly drop 10\% texts as well as 5\% point cloud renderings and camera embeddings for both ControlNet branches. To ensure generalization, the two ControlNets have only 20 layers while the main model has 40. Furthermore, we only inject the control signals from the ControlNets into the first 20 blocks of the main model. 
% As for inference, the classifier-free guidance scale is set to 5.0 for textual conditions, keeping other guidance on the default scale 1.0. 

\noindent\textbf{Datasets.} 
The training data of our camera control includes DL3DV~\cite{ling2024dl3dv}, Real10k~\cite{zhou2018Stereo}, Tartanair~\cite{tartanair2020iros}, Map-Free-Reloc~\cite{arnold2022map}, WildRGBD~\cite{xia2024rgbd}, and UE5 rendering data. For memory training, we discard Real10k due to its low-quality images and overly simple camera trajectories.
We further eliminate the Tartanair and narrow the frame interval of DL3DV for stable DMD training. More details are discussed in the supplementary.

\begin{table*}
\centering
\caption{\textbf{Quantitative results of OOD benchmark with WorldScore~\cite{duan2025worldscore} images}. $^*$ indicates the baseline version of our method without any memory mechanism, while the `full' version denotes adding both GGM and SSM. The `DMD' version is based on `WorldStereo-full'. Inference times are all tested with 8 H20 GPUs.
\protect\footnotemark
\label{tab:ood_exp3}}
\vspace{-0.125in}
\small
\setlength{\tabcolsep}{3pt}
\begin{tabular}{l|ccc|cccccc|c}
\toprule
 & RotErr$\downarrow$ & TransErr$\downarrow$ & ATE$\downarrow$ & Q-Align-I$\uparrow$ & Q-Align-V$\uparrow$ & CLIP-I$\uparrow$ & CLIP-T$\uparrow$ & CLIP-IQA+$\uparrow$ & Laion-Aes$\uparrow$ & Time (sec)\tabularnewline
\midrule
Voyager~\cite{huang2025voyager} & 0.678 & 0.630 & 1.343 & 3.279 & 0.664 & 82.63 & \textbf{11.399} & 0.414 & 4.994 & \textcolor{gray}{343} \tabularnewline
SEVA~\cite{zhou2025stable} & 0.171 & 0.540 & 1.023 & 3.907 & 0.782 & 91.74 & 11.194 & 0.514 & 5.331 & \textcolor{gray}{90} \tabularnewline
Gen3C~\cite{ren2025gen3c} & 0.220 & 0.275 & 1.071 & 4.094 & 0.820 & 91.78 & \underline{11.270} & 0.518 & 5.332 & \textcolor{gray}{158} \tabularnewline
Uni3C~\cite{cao2025uni3c} & 0.155 & 0.192 & 0.572 & 4.202 & 0.846 & 91.63 & 11.115 & 0.549 & 5.350 & 162 \tabularnewline
\midrule
WorldStereo* & \underline{0.132} & \underline{0.178} & \underline{0.542} & 4.273 & 0.860 & \underline{91.94} & 11.145 & 0.559 & 5.352 & 162 \tabularnewline
WorldStereo-GGM & \textbf{0.129} & \textbf{0.162} & 0.706 & \textbf{4.339} & \textbf{0.875} & \textbf{92.24} & 11.189 & \underline{0.572} & \textbf{5.408} & 162 \tabularnewline
WorldStereo-Full & 0.145 & 0.253 & 0.667 & 4.287 & 0.866 & 91.80 & 11.169 & 0.561 & 5.369 & 173 \tabularnewline
WorldStereo-DMD & 0.146 & 0.203 & \textbf{0.504} & \underline{4.338} & \underline{0.874} & 91.15 & 11.183 & \textbf{0.573} & \underline{5.393} & 9 \tabularnewline
\bottomrule
\end{tabular}
\vspace{-0.18in}
\end{table*}

\footnotetext{All methods are evaluated under their pre-defined resolutions and frame numbers, while the settings of Uni3C and WorldStereo series are the same for fairness (512p and 81-frame).}

\subsection{Results of Camera Control and Visual Quality}

\paragraph{Settings and Metrics.} To fairly evaluate the camera controllability of camera-guided models trained on different datasets, we introduce a new out-of-distribution (OOD) benchmark as shown in \Cref{tab:ood_exp3}. We select 100 images from the static subset of WorldScore~\cite{duan2025worldscore} as the initial frames of this benchmark, including high-quality samples across real-world, stylized, indoor, and outdoor scenarios.
Next, we randomly combine translation, rotation, and panning to construct complex camera trajectories.
To verify the camera precision, we employ WorldMirror~\cite{liu2025worldmirror} on the generated videos to extract predicted cameras and compare Rotation Error (RotErr), Translation Error (TransErr), and Absolute Trajectory Error (ATE) against ground-truth camera guidance. 
We further evaluate various quality assessments (Q-Align-Image\&Video~\cite{wu2023q}, CLIP-Image\&Text~\cite{radford2021learning}, CLIP-IQA+~\cite{wang2023exploring}, Laion-Aes~\cite{schuhmann2022improved}) via IQA-PyTorch~\cite{pyiqa} to verify the overall image and video quality.

\paragraph{Analysis.}
From \Cref{tab:ood_exp3}, the basic version WorldStereo$^*$ outperforms other competitors with superior camera control and visual quality. 
Moreover, we provide ablation studies on our memory components (GGM, SSM), where the memory bank and 3D cache only store the first frame's information. While the memory mechanism provides no benefit in single-image conditioned camera control, these results confirm that our memory-based training preserves the model's generalization and visual quality. In particular, GGM even improves the overall quality of the generated videos, while SSM slightly degrades performance, but enables strong fine-grained details recovery, as verified in \Cref{fig:ablation_study}(d). We also include the result of DMD, replacing the VDM backbone of our full model with the distilled DiT, showcasing impressive camera control and consistent quality, and a rapidly reduced inference cost.

\subsection{Single-View Reconstruction Benchmark}
\label{sec:3d_world}

\begin{table}
\centering
\caption{\textbf{Quantitative results of 3D reconstruction} based on Tanks-and-Temples~\cite{knapitsch2017tanks} and MipNeRF360~\cite{barron2022mip}. 
$^*$ indicates the baseline version of our method without any memory mechanism.}
\label{tab:rec_exp2}
\vspace{-0.125in}
\footnotesize
\setlength{\tabcolsep}{0pt} % Set column separation to 0, as \extracolsep will manage it.

% Use tabular* to set the table width to the text width
% @{\extracolsep{\fill}} distributes the extra space between columns
\begin{tabular*}{\linewidth}{@{\extracolsep{\fill}} llccccc} 
\toprule
 & \textbf{Method} & F1-Score~$\uparrow$ & AUC~$\uparrow$ & RotErr~$\downarrow$ & TransErr~$\downarrow$ & ATE~$\downarrow$ \\
\midrule
% Use \multirow to merge rows for the 'Tanks-and-Temples' dataset
\multirow{9}{*}{\rotatebox{90}{\textbf{Tanks\&Temples}}} & Uni3C~\cite{cao2025uni3c} & 0.424 & 0.378 & 0.362 & 0.1017 & 0.1572\\
& Gen3C~\cite{ren2025gen3c} & 0.416 & 0.380 & 0.342 & {0.0949} & 0.1704\\
& SEVA~\cite{zhou2025stable} & 0.286 & 0.293 & 0.379 & {0.0949} & 0.1815 \\
& Lyra~\cite{bahmani2025lyra} & 0.227 & 0.193  & -- & -- & -- \\
& VMem~\cite{li2025vmem} & 0.386 & 0.375 & 0.533 & 0.1510 & 0.1922 \\
& WorldStereo$^*$ & 0.447 & 0.389 & 0.377 & 0.0990 & {0.1545} \\
& WorldStereo-GGM & 0.485 & \underline{0.411} & \textbf{0.224} & \textbf{0.0885} & \textbf{0.1350} \\
& WorldStereo-Full & \textbf{0.578} & \textbf{0.437} & \underline{0.247} & \underline{0.0927} & \underline{0.1501} \\
& WorldStereo-DMD & \underline{0.534} & {0.410} & 0.291 & 0.1001 & 0.1547 \\
\midrule % Separator line between the two datasets
% Use \multirow to merge rows for the 'MipNeRF360' dataset
\multirow{9}{*}{\rotatebox{90}{\textbf{MipNeRF360}}} & Uni3C~\cite{cao2025uni3c} & 0.352 & 0.347  & 0.112 & 0.0086 & \underline{0.0104} \\
& Gen3C~\cite{ren2025gen3c} & 0.356 & 0.340 & 0.349 & 0.0220 & 0.0318\\
& SEVA~\cite{zhou2025stable} & 0.332 & 0.311 & 0.282 & 0.0138 & 0.0295 \\
& Lyra~\cite{bahmani2025lyra} & 0.203 & 0.263  & -- & -- & -- \\
& VMem~\cite{li2025vmem} & 0.256 & 0.245 & 0.403 & 0.0392 & 0.0752 \\
& WorldStereo$^*$ & 0.350 & 0.342  & \textbf{0.097} & \textbf{0.0076} & \textbf{0.0099} \\
& WorldStereo-GGM & 0.342 & 0.346 & \underline{0.107} & \underline{0.0079} & 0.0206 \\
& WorldStereo-Full & \textbf{0.406} & \textbf{0.402} & 0.114 & {0.0080} & 0.0132 \\
& WorldStereo-DMD & \underline{0.390} & \underline{0.387} & 0.159 & 0.0106 & 0.0267 \\
\bottomrule
\end{tabular*}
\vspace{-0.23in}
\end{table}

\paragraph{Settings and Metrics.}
To comprehensively evaluate the quality of video generation and multi-trajectory consistency for reconstruction, we propose a novel 3D reconstruction benchmark based on single-view generation.
Our benchmark comprises the training split of Tanks-and-Temples~\cite{knapitsch2017tanks} and MipNeRF360~\cite{barron2022mip}.
The training split of Tanks-and-Temples has ground-truth point clouds.
For MipNeRF360, we first reconstruct the global point clouds scene by MVS~\cite{cao2024mvsformer++} and then crop the centric foreground areas as pseudo ground-truth point clouds. Each scene of both datasets only provides a single image as the initial frame.
More details about the data curation of this benchmark are discussed in the supplementary.
The evaluation workflow includes: 1) generating videos along 4 pre-defined trajectories of up, left, right rotations, and orbit as shown in \Cref{fig:3d_vis}(a); 2) reconstructing point clouds through WorldMirror~\cite{liu2025worldmirror}; 3) aligning the reconstructed point clouds to the ground-truth ones.
For MipNeRF360, we can leverage the initial frame's MVS depth as an anchor to align point-to-point matched WorldMirror points via Umeyama translation~\cite{umeyama1991least}.
For Tanks-and-Temples, we first align the rotation and translation via the first frame's camera. Then, we apply ICP~\cite{besl1992method} to optimize a refined scale between the reconstructed and ground-truth point clouds.
Except for camera metrics, we include two widely used point cloud metrics. 
Specifically, we get the precision and recall of point cloud within scene-wise distance thresholds manually adjusted as~\cite{knapitsch2017tanks}, which evaluate the accuracy and completeness of reconstructed point clouds.
Besides, we incorporate the point cloud AUC, calculated as the area under the ROC curve of precision and recall with varying thresholds.

\paragraph{Analysis.}
From \Cref{tab:rec_exp2}, WorldStereo$^*$ without any memory mechanisms already outperforms other methods, while our full model, enhanced with GGM and SSM, achieves substantial improvements on both reconstruction and camera precision.
Moreover, WorldStereo-DMD still performs well in 3D reconstruction, showing impressive consistency with significant acceleration. 

\subsection{Extensive Qualitative Results}
\label{sec:qualitative}

\paragraph{Qualitative Comparison.}

\begin{figure*}
    \centering
    \includegraphics[width=1.0\linewidth]{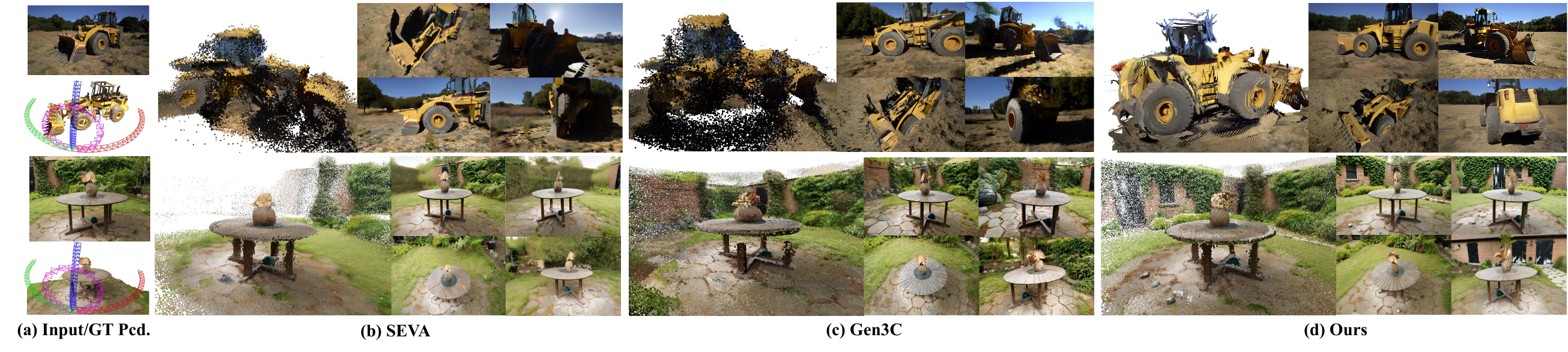}
    \vspace{-0.3in}
    \caption{\textbf{Results of 3D reconstruction benchmark.} The column (a) shows input views and ground-truth point clouds with pre-defined four trajectories (up, left, right rotations, and orbit). We compare the qualitative results of reconstructed point clouds (left) and generated novel views (right) for each method. \label{fig:3d_vis}}
    \vspace{-0.1in}
\end{figure*}

We show qualitative results in \Cref{fig:3d_vis}. As shown in \Cref{fig:3d_vis}(a), our 3D reconstruction benchmark features high-quality ground-truth point clouds that focus on foreground objects. The compared methods must retain consistent outcomes with symmetrical and logically coherent structures to achieve good point cloud scores. Thus, this benchmark serves as a reliable evaluation for 3D scene generation.
SEVA~\cite{zhou2025stable} suffers from distorted structures and blurry backgrounds, while Gen3C~\cite{ren2025gen3c} faces challenges in producing consistent videos under different trajectories---both methods yield ambiguous, incomplete reconstructions.
In contrast, point clouds reconstructed from WorldStereo's outputs enjoy remarkable completeness and precision, while its novel views remain consistent across different trajectories and exhibit superior quality.

\paragraph{Memory Mechanisms.}

\begin{figure*}
    \centering
    \includegraphics[width=1.0\linewidth]{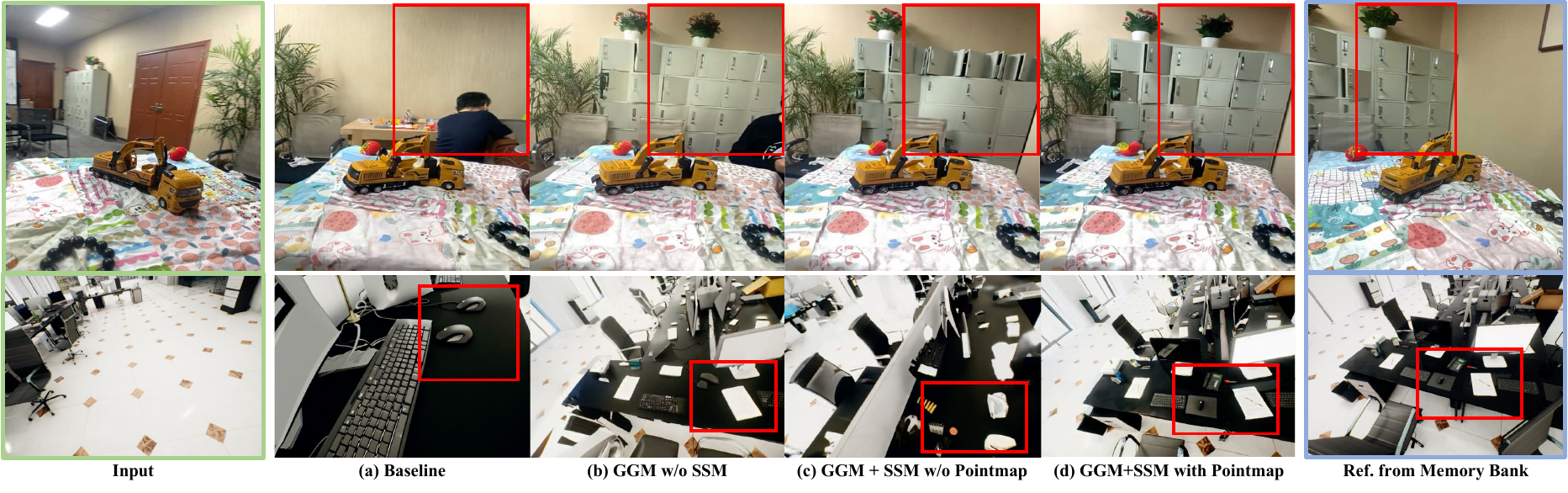}
    \vspace{-0.25in}
    \caption{\textbf{Ablation studies of memory components.} 
    Please see the red-framed regions to check the consistency compared to retrieved references. Baseline results are generated without any memory.
    GGM can capture coarse structures, but loses fine-grained details. Moreover, the incorporation of pointmap significantly enhances the consistency gained via the reference frames retrieved from the memory bank.\label{fig:ablation_study}}
    \vspace{-0.15in}
\end{figure*}

We show the qualitative ablation studies in \Cref{fig:ablation_study} to verify the individual effectiveness of each memory component. The baseline without any memory mechanism randomly hallucinates new objects in novel views, while GGM largely retains the structural consistency and improves camera control for disparate viewpoint changes (second row), benefiting from the incrementally updated 3D cache.
The ability of SSM to preserve fine-grained consistency with the memory bank reference stems from its attention mechanism, which leverages images with 3D guidance instead of drawing information solely from coarse-grained point clouds.
% Moreover, SSM preserves the fine-grained consistency compared to the reference retrieved from the memory bank. 
We clarify that the 3D correspondence-based pointmap guidance is critical for the SSM to focus on correct matching regions.

% We show the qualitative ablation studies in \Cref{fig:ablation_study} to verify the effectiveness of each memory component. TThe baseline without memory randomly hallucinates new objects in novel views, In contrast, GGM provides a consistent geometric scaffold via its incrementally updated 3D cache, thus retaining structural integrity and improving camera control across disparate viewpoints (second row). Furthermore, SSM preserves fine-grained appearance consistency. We clarify that the 3D correspondence-based pointmap guidance is critical for this, as it enables the SSM to accurately map textures to the correct geometric regions, preventing visual artifacts.

\paragraph{3D Panorama Generation.}

\begin{figure}
    \centering
    \includegraphics[width=1.0\linewidth]{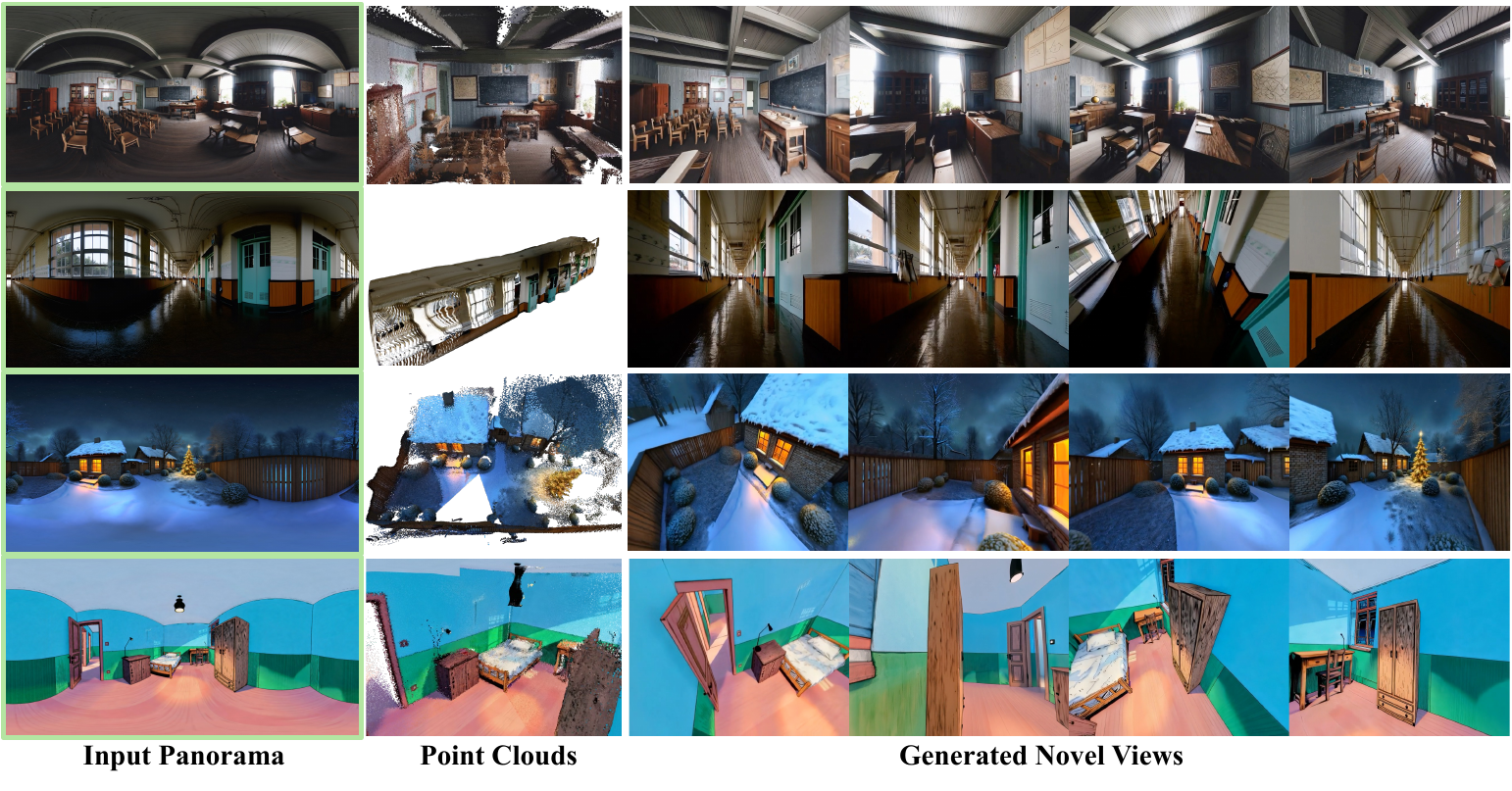}
    \vspace{-0.25in}
    \caption{\textbf{Results of 3D panorama generation.} Please zoom-in for more details of reconstructed point clouds and novel views.\label{fig:panorama}}
    \vspace{-0.2in}
\end{figure}

Thanks to memory capabilities, WorldStereo can be easily extended to 3D panorama generation as shown in \Cref{fig:panorama}. Different from panoramic video generation~\cite{yin2025panoworld,yang2025matrix,xia2025panowan}, our method largely preserves the generalization of the foundational VDM while producing high-resolution perspective views (all panorama inferences are performed at 576p). Both aforementioned points are critical for 3D reconstruction.
Formally, we split the panorama into 27 frames under FoV $90\times120$ as the initial memory bank, and leverage the panoramic depth estimation of MoGe~\cite{wang2025moge} to build the 3D cache. We heuristically design wonder trajectories as used in our reconstruction benchmark (\Cref{sec:3d_world}) and generate videos for the middle three non-overlap images via WorldStereo.

\section{Conclusion}

In this paper, we present WorldStereo, a novel camera-guided video generation framework tailored for 3D reconstruction. To address the challenge of capturing long-sequence viewpoints while maintaining 3D consistency, WorldStereo generates multiple consistent videos via two memory mechanisms: Global-Geometry Memory (GGM) and Spatial-Stereo Memory (SSM). GGM incrementally updates a point cloud-based 3D cache to enhance the coherent 3D structures, while SSM learns coherence between generated and retrieved views through 3D correspondence to preserve fine-grained details. 
Moreover, we modify the distribution matching distillation (DMD) strategy to accelerate WorldStereo for fast inference with negligible performance drop.
Additionally, we develop a new 3D reconstruction benchmark for evaluating video generation performance. 
WorldStereo demonstrates strong performance and generalizes effectively to diverse tasks, such as object-centric, face-forward, and 3D panorama generation.

{
    \small
    \bibliographystyle{ieeenat_fullname}
    \bibliography{main}
}

% \input{supp}

\appendix

%%%%%%%%% BODY TEXT - ENTER YOUR RESPONSE BELOW
\section{Datasets}

We summarize the datasets of this work in \Cref{tab:dataset_details}, including both training and evaluation dataset settings. Note that most of our training data is publicly available.
We randomly sample a subset for each data for each epoch, according to its diversity and video, trajectory quality.

\begin{table}[h]
\centering
\caption{\textbf{Dataset details of WorldStereo.} 
The training datasets used for various training processes include DL3DV~\cite{ling2024dl3dv}, Re10K~\cite{zhou2018Stereo}, Tartainair~\cite{tartanair2020iros}, Map-Free-Reloc~\cite{arnold2022map}, WildRGBD~\cite{xia2024rgbd}, and UCo3D~\cite{liu2025uncommon}.
We dynamically sample subsets for each dataset across training epochs.\label{tab:dataset_details}}
\vspace{-0.1in}
{\footnotesize{}}%
\setlength{\tabcolsep}{2.2pt} % 减小列间距
\resizebox{\linewidth}{!}{
\begin{tabular}{lccccccc}
\toprule  
 & \multicolumn{4}{c}{{\footnotesize{}Train Processes}} & \multicolumn{2}{c}{{\footnotesize{}Scene Number}}\tabularnewline
 & {{\footnotesize{}Camera}} & {{\footnotesize{}GGM}} & {{\footnotesize{}SSM}} & {{\footnotesize{}DMD}} & {\footnotesize{}Total} & {\footnotesize{}Epoch}\tabularnewline
\midrule 
{\footnotesize{}DL3DV} & {\footnotesize{}$\checkmark$} & {\footnotesize{}$\checkmark$} & {\footnotesize{}$\checkmark$} &{\footnotesize{}$\checkmark$}  & {\footnotesize{}9,174} & {\footnotesize{}19,868}\tabularnewline
{\footnotesize{}Re10K} & {\footnotesize{}$\checkmark$} & {\footnotesize{}$\checkmark$} &  &  & {\footnotesize{}26,544} & {\footnotesize{}12,000}\tabularnewline
{\footnotesize{}Tartainair} & {\footnotesize{}$\checkmark$} & {\footnotesize{}$\checkmark$} & {\footnotesize{}$\checkmark$} & {\footnotesize{}$\checkmark$} & {\footnotesize{}2,214}  & {\footnotesize{}2,142}\tabularnewline
{\footnotesize{}Map-Free-Reloc} & {\footnotesize{}$\checkmark$} & {\footnotesize{}$\checkmark$} & {\footnotesize{}$\checkmark$} & {\footnotesize{}$\checkmark$} & {\footnotesize{}960} & {\footnotesize{}910}\tabularnewline
{\footnotesize{}WildRGBD} & {\footnotesize{}$\checkmark$} & {\footnotesize{}$\checkmark$} & {\footnotesize{}$\checkmark$} & {\footnotesize{}$\checkmark$} & {\footnotesize{}23,005} & {\footnotesize{}3,000}\tabularnewline
{\footnotesize{}UCo3D} & {\footnotesize{}$\checkmark$} &  &  & {\footnotesize{}$\checkmark$} & {\footnotesize{}164,455} & {\footnotesize{}10,000}\tabularnewline
{\footnotesize{}UE-Render} & {\footnotesize{}$\checkmark$} & {\footnotesize{}$\checkmark$} & {\footnotesize{}$\checkmark$} & {\footnotesize{}$\checkmark$} & {\footnotesize{}5,939} & {\footnotesize{}4,000}\tabularnewline
\bottomrule 
\end{tabular}{\footnotesize\par}
}
\end{table}

\section{More Quantitative Ablation Studies}
\paragraph{Benchmark for Memory Components.}
To quantitatively assess the contribution of various modules for video generation and memory capabilities, we have established a benchmark consisting of 100 scenes drawn from our diverse test set in \Cref{tab:membench}, containing data from DL3DV~\cite{ling2024dl3dv}, Map-Free-Reloc~\cite{arnold2022map}, WildRGBD~\cite{xia2024rgbd}, Tartanair~\cite{tartanair2020iros}, and UE5-rendered scenes. It features a wide array of challenging situations, including real and virtual environments, indoor and outdoor settings, and varying complexities of camera motion. 
The construction of the memory bank involves selecting videos that have a temporal overlap of 30\% to 90\%. To increase the diversity and complexity of test samples, a random dropping strategy is applied to the reference frames. Specifically, there is a 10\% chance that all reference frames are dropped, yielding an empty memory bank, and a 10\% chance that no frames are dropped. For the remaining scenarios, each reference frame associated with a target view is randomly dropped with a probability of 40\%.
Since this benchmark contains ground-truth video views to evaluate the image fidelity, we can include PSNR, SSIM, and LPIPS to verify the effectiveness of memory components.
% We employ camera metrics such as Rotational Error (RotErr), Translational Error (TransErr), and Absolute Trajectory Error (ATE) for evaluating camera pose accuracy, and PSNR for image fidelity. 
% As shown in \Cref{tab:membench}, the results are\TODO{...}

\begin{table}[h]
\centering
\caption{\textbf{Quantitative ablation for memory components.}
$^*$ indicates our baseline without any memory mechanism, while the `full' version denotes adding both GGM and SSM. The `DMD' version is based on `WorldStereo-full'.
\label{tab:membench}}
\vspace{-0.125in}

\resizebox{\linewidth}{!}{%
\begin{tabular}{l|ccc|ccc}
\toprule
 & RotErr$\downarrow$ & TransErr$\downarrow$ & ATE$\downarrow$ & PSNR$\uparrow$ & SSIM$\uparrow$ & LPIPS$\downarrow$  \tabularnewline
\midrule
% WorldStereo* & 14.64 & 0.443 & 0.412 & 1.300 & 0.112 & 0.237 \tabularnewline
% WorldStereo-GGM  & 17.45 & 0.532 & 0.288 & 0.699 & 0.067 & 0.131 \tabularnewline
% WorldStereo-Full & 18.40 & 0.561 & 0.283 & 0.748 & 0.079 & 0.142 \tabularnewline
% WorldStereo-DMD &  &  &  &  &  &  \tabularnewline
% \bottomrule
WorldStereo* & 1.300 & 0.112 & 0.237 & 14.64 & 0.443 & 0.412 \tabularnewline
WorldStereo-GGM  & \textbf{0.699} & \textbf{0.067} & \underline{0.131} & 17.45 & 0.532 & \underline{0.288} \tabularnewline
WorldStereo-Full & \underline{0.748} & 0.079 & 0.142 & \textbf{18.40} & \textbf{0.561} & \textbf{0.283} \tabularnewline
WorldStereo-DMD & 0.772 & \underline{0.069} & \textbf{0.130} & \underline{18.04} & \underline{0.546} & 0.289 \tabularnewline
\bottomrule
\end{tabular}%
}
\vspace{-0.18in}
\end{table}

\paragraph{Camera Control Evaluation.}
We build a test set (DL3DV, Real10k~\cite{zhou2018Stereo}, Map-Free-Reloc, WildRGBD, Tartanair, and UCo3D~\cite{liu2025uncommon}) to evaluate the capability of camera control. 
Different from the aforementioned memory ablation benchmark, all samples in this benchmark only contain depth and point clouds extracted from the first frame without additional information from other views.
A comparative analysis was conducted among Uni3C~\cite{cao2025uni3c}, WorldStereo, and its DMD-accelerated variant, WorldStereo-DMD. 
We did not apply memory components in this part.
The primary metrics for comparison are camera trajectory accuracy and resulting image quality. The results presented in \Cref{tab:cambench}, indicate that our proposed method yields substantial gains in camera control precision compared to Uni3C, while also offering improvements to video quality. Notably, the application of DMD acceleration results in a 20x increase in inference speed for the WorldStereo-DMD model, with no significant drop in qualitative output and camera control.

% \begin{table}
% \centering
% \caption{\textbf{Quantitative results of in-domain test set}. 
% \protect\footnotemark
% \label{tab:indomain}}
% \vspace{-0.125in}
% \small
% \setlength{\tabcolsep}{3pt}
% \begin{tabular}{l|ccc|ccc}
% \toprule
%  & RotErr$\downarrow$ & TransErr$\downarrow$ & ATE$\downarrow$ & PSNR$\uparrow$ & SSIM$\uparrow$ & LPIPS$\downarrow$ 
% \midrule
% Uni3C  &  &  &  &  &  &  \tabularnewline
% WorldStereo &  &  &  &  &  &  \tabularnewline
% % WorldStereo-Full &  &  &  &  &  &  \tabularnewline
% WorldStereo-DMD &  &  &  &  &  &  \tabularnewline
% \bottomrule
% \end{tabular}
% \vspace{-0.18in}
% \end{table}

\begin{table}[h]
\centering
\caption{\textbf{Quantitative results for camera control.} 
\label{tab:cambench}}
\vspace{-0.125in}
\resizebox{\linewidth}{!}{%
\begin{tabular}{l|ccc|ccc}
\toprule
 & RotErr$\downarrow$ & TransErr$\downarrow$ & ATE$\downarrow$ & PSNR$\uparrow$ & SSIM$\uparrow$ & LPIPS$\downarrow$  \tabularnewline
\midrule
Uni3C  & 1.816 & 0.122 & 0.359 & \textbf{14.28} & \textbf{0.475} & \underline{0.449} \tabularnewline
WorldStereo & \textbf{1.384} & \underline{0.115} & \underline{0.347} & \underline{14.27} & \underline{0.473} & \textbf{0.441} \tabularnewline
% WorldStereo-Full &  &  &  &  &  &  \tabularnewline
WorldStereo-DMD & \underline{1.435} & \textbf{0.085} & \textbf{0.211} & 13.78 & 0.436 & 0.463 \tabularnewline
\bottomrule
\end{tabular}%
}
\vspace{-0.18in}
\end{table}

\begin{figure*}
    \centering
    \includegraphics[width=1.0\linewidth]{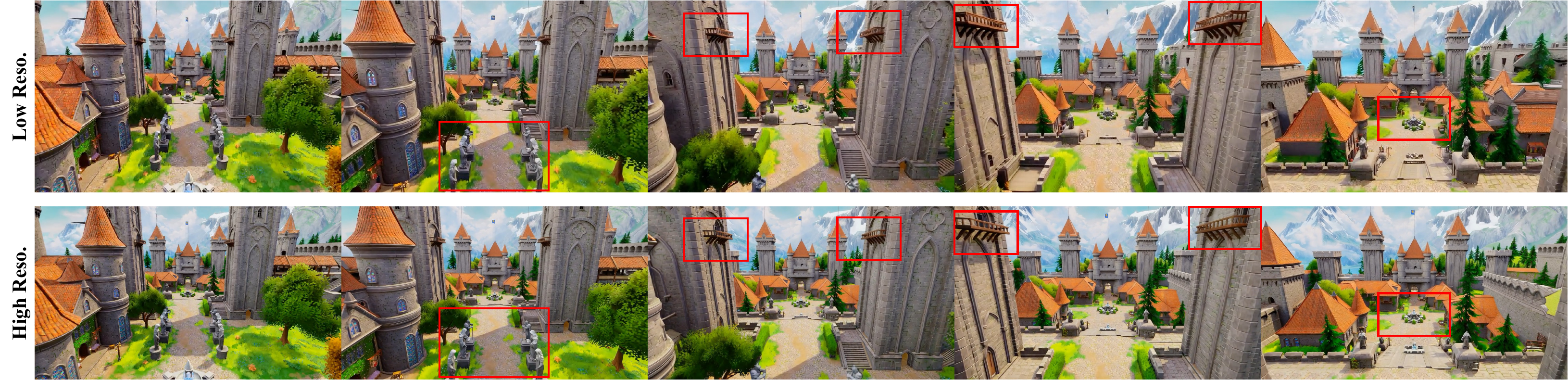}
    \vspace{-0.2in}
    \caption{\textbf{Qualitative Results of different resolutions.} High-resolution (720$\times$1280) yields sharper images with richer details than the low-resolution ones (480$\times$768), as demonstrated in the red-boxed regions. \label{fig:reso}}
    % \vspace{-0.2in}
\end{figure*}

\paragraph{High-Resolution Inference.}
As presented in Section 4.1, our model is capable of performing inference on high-resolution $720 \times 1280$ videos, despite only training exclusively on 480p data. This capability highlights the success of our method in retaining the generalization ability of the base model, which is inherently capable of processing 720p videos. Furthermore, as shown in \Cref{fig:reso}, the model can generate images with enhanced detail by directly performing inference at high resolutions, obviating the need for retraining. Nevertheless, to balance performance with computational efficiency, the experiments in this paper are carried out using 480p resolution.
% \Cref{tab:resolution}, inference at higher resolution results in a significant enhancement of image quality, as reflected by a marked improvement in reconstruction metrics. Simultaneously, the camera metrics of high resolution version are comparable to the 480p version. 
% These evaluations are conducted on an in-domain test set which is different from \TODO{}

% \begin{table}[h]
% \centering
% \caption{\textbf{Quantitative results of different resolutions}. 
% \label{tab:resolution}}
% \vspace{-0.125in}

% \resizebox{\linewidth}{!}{%
% \begin{tabular}{l|ccc|ccc}
% \toprule
%  & RotErr$\downarrow$ & TransErr$\downarrow$ & ATE$\downarrow$ & PSNR$\uparrow$ & SSIM$\uparrow$ & LPIPS$\downarrow$  \tabularnewline
% \midrule
% 480 \times 768   & \textbf{0.303} & \textbf{0.0168} & 0.0676 & 16.636 & 0.571 & 0.382 \tabularnewline
% 720 \times 1280  & 0.332 & 0.0192 & \textbf{0.0673} & \textbf{18.357} & \textbf{0.691} & \textbf{0.347} \tabularnewline
% \bottomrule
% \end{tabular}%
% }
% \end{table}

% \section{Details of 3D Reconstruction Benchmark}
\section{Trajectory Settings}

\begin{table}[h]
\centering
\caption{\textbf{Overlapping FoV scores of 3D panorama generation (trajectory order ablation)}. Memory bank settings: `only panorama' uses 24 views split from the panorama; others are incrementally updated with generations from different trajectory orders. Higher scores mean that more relevant frames are retrieved.
`reference prop.' indicates the proportion of retrieved frames belonging to panoramic (pano.) or generated (gen.) frames.
\label{tab:traj_ablation}}
\vspace{-0.125in}
\footnotesize
\setlength{\tabcolsep}{2.5pt}
\begin{tabular}{c|ccccc|cc}
\toprule
\multirow{2}{*}{} & \multirow{2}{*}{orbit} & \multirow{2}{*}{up} & \multirow{2}{*}{right} & \multirow{2}{*}{left} & \multirow{2}{*}{all} & \multicolumn{2}{c}{reference prop.}\tabularnewline
 &  &  &  &  &  & pano. & gen.\tabularnewline
\midrule
only panorama & 46.0 & 43.4 & 38.4 & 38.5 & 167.2 & 100.0\% & 0.00\%\tabularnewline
up$\rightarrow$right$\rightarrow$left$\rightarrow$orbit & 50.7 & 43.4 & 39.6 & 41.3 & 175.0 & 65.17\% & 34.83\%\tabularnewline
orbit$\rightarrow$up$\rightarrow$right$\rightarrow$left & 46.9 & 46.2 & 41.3 & 42.8 & 177.1 & 54.89\% & 45.11\%\tabularnewline
orbit$\rightarrow$right$\rightarrow$left$\rightarrow$up & 46.9 & 46.9 & 41.1 & 42.8 & 177.6 & 53.66\% & 46.34\%\tabularnewline
up$\rightarrow$orbit$\rightarrow$right$\rightarrow$left & 49.2 & 43.4 & 41.3 & 42.8 & 176.7 & 55.55\% & 44.45\%\tabularnewline
right$\rightarrow$left$\rightarrow$up$\rightarrow$orbit & 50.7 & 45.2 & 38.4 & 41.2 & 175.6 & 60.89\% & 39.11\%\tabularnewline
right$\rightarrow$left$\rightarrow$orbit$\rightarrow$up & 49.3 & 46.9 & 38.4 & 41.2 & 176.3 & 62.43\% & 37.57\%\tabularnewline
\bottomrule
\end{tabular}
\end{table}

\begin{figure}[h]
    \vspace{-0.35in}
    \centering
    \includegraphics[width=1.0\linewidth]{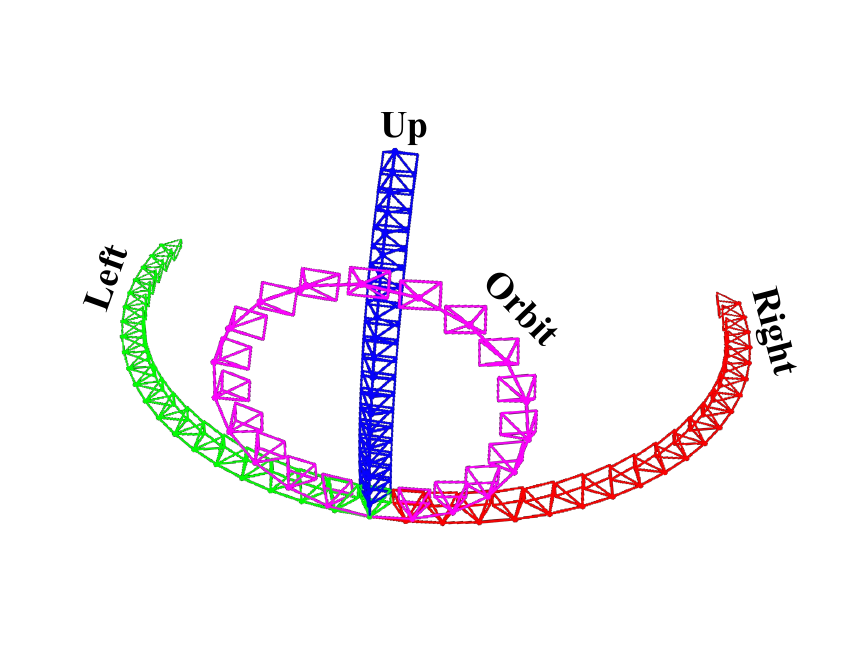}
    \vspace{-0.5in}
    \caption{\textbf{Illustration of the trajectory.} \label{fig:traj}}
    \vspace{-0.1in}
\end{figure}

To determine an optimal trajectory sequence for high-quality 3D reconstruction, we performed a series of ablation studies across 10 panoramic scenes, with results summarized in \Cref{tab:traj_ablation}.
The memory bank was initialized with 24 views extracted from the panorama, while others are incrementally updated with generations from different trajectory orders (see ~\Cref{fig:traj}).
From \Cref{tab:traj_ablation}, the updated memory bank exhibits more reliable references compared to the baseline setting that solely relies on panoramic images. 
Notably, the orbit trajectory---with its information-rich viewing angles, should be prioritized as the initial trajectory.
We finally use `orbit$\rightarrow$up$\rightarrow$right$\rightarrow$left' as our default trajectory sequence. 
This ordering is justified by the fact that left and right rotations are more critical to 3D reconstruction than other trajectories. Positioning these rotations last allows them to leverage a larger number of pre-accumulated memory frames, thereby enhancing their contribution to the final reconstruction.

Specifically, we set the rotation angles for the upward, leftward, and rightward rotations to be 45$^\circ$, 90$^\circ$, 90$^\circ$, respectively. For these rotational movements, the distance from the rotation center is configured as the median depth of the scene, while the radius of the orbital trajectory is set to 0.3 times the median depth. For some face-forwarding scenes (\eg, `room' in MipNeRF360~\cite{barron2022mip}), we manually reduce the rotation angles to avoid the collision.

\section{Details of Data Curation}

\paragraph{Sampling Strategy for Training GGM.}
A global point cloud is constructed from multi-view images and their associated depth maps. Our GGM training process begins with the point cloud of the initial frame, $X_{pcd}$. We then augment this by randomly sampling 1 to 4 additional frames, generating their respective point clouds ($\hat{X}_{pcd}$) from depth information, and aligning them to the coordinate system of $X_{pcd}$. To mitigate overfitting, we apply two data augmentation techniques to $\hat{X}_{pcd}$. The first one is random masking, which nullifies 30\% to 70\% of randomly selected pixels in the depth map. The second one is contiguous masking, which applies a randomly positioned rectangular mask that occludes 20\% to 70\% of the depth map's area. These augmentations allow the GGM module to effectively learn global geometric features without being misled by spurious or incorrect global point cloud conditions.

\paragraph{Sampling Strategy for Training SSM.}
Following the procedure for SSM training outlined in Section 3.3, we construct training pairs from our multi-view dataset, ensuring each reference-target pair has a temporal overlap of 30\% to 90\% and the same number of frames. To enhance model robustness, we employ a reference dropout strategy. Specifically, for any given training sample, the entire reference condition is omitted with a 10\% probability. Otherwise, for each target frame, its corresponding reference frame is randomly dropped with a 30\% probability, thereby nullifying the reference condition for that frame. This process generates an unordered reference set for SSM training, which is closer to real-world application scenarios. Consequently, instead of encoding them as a coherent video, we process each remaining reference frame independently, encoding it into a latent representation using a pre-trained VAE~\cite{wang2025wan}.

% \paragraph{Sampling strategy for DMD.}

\end{document}